\documentclass{article}

    \usepackage[preprint,nonatbib]{neurips_2023}

\usepackage[utf8]{inputenc} %
\usepackage[T1]{fontenc}    %
\usepackage[pagebackref,breaklinks,colorlinks]{hyperref}
\usepackage{url}            %
\usepackage{booktabs}       %
\usepackage{amsfonts}       %
\usepackage{nicefrac}       %
\usepackage{microtype}      %
\usepackage{xcolor}         %
\usepackage{xspace}
\usepackage{amsmath}
\usepackage{multirow}
\usepackage{booktabs}
\usepackage{graphicx}  %
\usepackage{amssymb}   %
\usepackage{tabularx}
\usepackage{adjustbox}
\usepackage{units}
\usepackage{mathtools}
\usepackage{caption}
\usepackage{arydshln}
\usepackage{amssymb}%
\usepackage{pifont}%
\usepackage{comment}
\usepackage{float}
\usepackage{titlesec} %
\newcommand{\cmark}{\ding{51}}%
\newcommand{\xmark}{\ding{55}}%

\DeclarePairedDelimiter{\norm}{\lVert}{\rVert}
\definecolor{DarkGreen}{rgb}{0.0, 0.7, 0.5}

\newcommand{\ie}{\emph{i.e.}\xspace}

\newcommand{\etal}{\emph{et al.}\xspace}
\newcommand{\neuralname}{NRMs\xspace}

\title{Enhancing Neural Rendering Methods\\with Image Augmentations}

\author{%
  Juan~C. Pérez\\
  KAUST\\
  \And
  Sara Rojas\\
  KAUST\\
  \And
  Jesus Zarzar\\
  KAUST\\
  \And
  Bernard Ghanem\\
  KAUST
}

\begin{document}

\maketitle

\begin{abstract}
Faithfully reconstructing 3D geometry and generating novel views of scenes are critical tasks in 3D computer vision.
Despite the widespread use of image augmentations across computer vision applications, their potential remains underexplored when learning neural rendering methods~(\neuralname) for 3D scenes.
This paper presents a comprehensive analysis of the use of image augmentations in \neuralname, where we explore different augmentation strategies. 
We found that introducing image augmentations during training presents challenges such as geometric and photometric inconsistencies for learning \neuralname from images.
Specifically, geometric inconsistencies arise from alterations in shapes, positions, and orientations from the augmentations, disrupting spatial cues necessary for accurate 3D reconstruction.
On the other hand, photometric inconsistencies arise from changes in pixel intensities introduced by the augmentations, affecting the ability to capture the underlying 3D structures of the scene. 
We alleviate these issues by focusing on color manipulations and introducing learnable appearance embeddings that allow \neuralname to explain away photometric variations.
Our experiments demonstrate the benefits of incorporating augmentations when learning \neuralname, including improved photometric quality and surface reconstruction, as well as enhanced robustness against data quality issues, such as reduced training data and image degradations. 
\end{abstract}

\section{Introduction}
Reconstructing 3D geometry and generating novel views are crucial tasks in 3D computer vision, with applications ranging from robotics to virtual and augmented reality, autonomous navigation, and digital content creation~\cite{carranza2003free, szeliski2022computer, snavely2006photo}.
Neural rendering methods (\neuralname) have gained traction as powerful approaches for learning 3D scene representations, capable of capturing complex geometry and appearance properties of diverse objects~\cite{tewari2022advances}.
As such, \neuralname are a powerful and versatile 3D representation that can be leveraged for other computer vision tasks~\cite{poole2022dreamfusion}.

Computer vision researchers customarily use augmentations to improve performance and robustness across tasks~\cite{alomar2023data}.
Specifically, a wide variety of tasks such as object recognition~\cite{resnet, simonyan2014very, perez2021enhancing}, object detection~\cite{li2017fully} and segmentation~\cite{moshkov2020test} draw sizable benefits from leveraging augmentations.
Interestingly, despite \neuralname being widely used in computer vision, they are yet to successfully exploit the power of image augmentations~\cite{wang2023benchmarking}.
We double down on this observation, and raise two pertinent questions:~(1)~how can image augmentations be used for learning \neuralname? and~(2)~how are \neuralname affected by these augmentations?

\begin{figure}[t] %
 \centering
 \includegraphics[width=1.\textwidth]{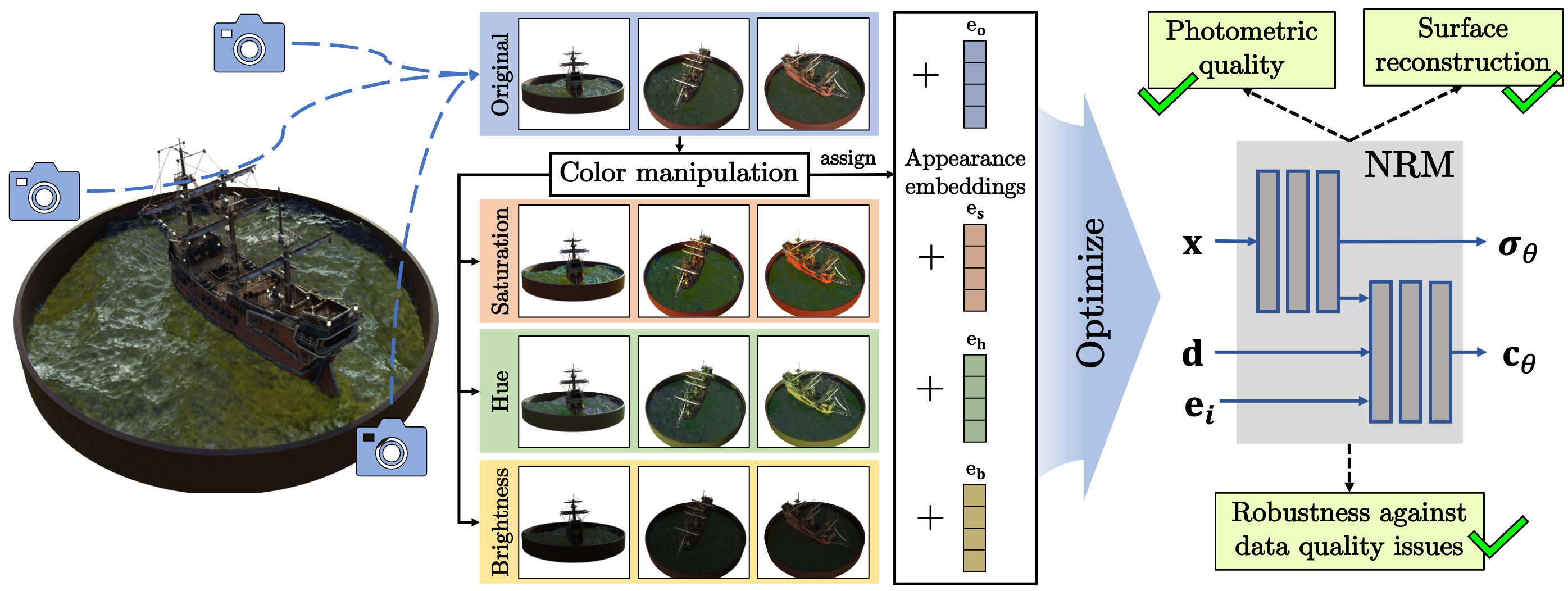}
 \caption{
     \textbf{Studying the impact of image augmentations on Neural Rendering Methods.}
     Introducing image augmentations into the training set of \neuralname presents geometric and photometric inconsistencies for optimization.
     We alleviate these issues, respectively, by~(1)~augmenting images only with color manipulations, and~(2)~introducing learnable appearance embeddings for each manipulation.
     These solutions allow us to study how \neuralname are affected by augmentations by directly training them on augmented datasets.
     Our study finds that such augmentations consistently improve photometric quality, surface reconstruction, and robustness against issues in data quality.
     Please refer to Section~\ref{sec:methodology} for specifics into how this pipeline is specialized for \textit{Static} and \textit{Dynamic} Image Augmentations (\ie SIA and DIA).
 }
 \label{fig:pull_fig}
\end{figure}

In this paper, we present an analysis of the use of image augmentations in \neuralname, focusing on two distinct setups: Static Image Augmentations (SIA) and Dynamic Image Augmentations (DIA).
In SIA, all training images are transformed with some fixed parameters and stored, effectively augmenting the training set.
In DIA, a transformation and its parameters are randomly sampled on-the-fly from a distribution at each training iteration.
Introducing either type of augmentation at training time presents two problems for \neuralname:~(1)~geometric inconsistencies, such as those caused by cropping and flipping, and~(2)~photometric inconsistencies, like those resulting from contrast enhancement.
Correspondingly, we address these issues by~(1)~exclusively using color manipulations, and~(2)~coupling each manipulation with learnable appearance embeddings that allow \neuralname to explain away photometric inconsistencies.
With these solutions in hand, we turn to evaluate the impact of image augmentations on various \neuralname, such as NeRF~\cite{mildenhall2021nerf}, NGP~\cite{mueller2022instant}, and NeuS~\cite{wang2021neus}.
Please refer to Figure~\ref{fig:pull_fig} for an overview of our approach and the main findings of our study.
Our experiments reveal that standard data augmentation techniques, known to enhance performance and robustness in computer vision models, indeed contribute analogously to the performance of \neuralname.
Furthermore, we find that the inexpensive SIA, despite its simplicity, provides larger and more consistent improvements than DIA both in terms of photometric quality and reconstruction.
We further investigate the SIA setup, and find that it provides benefits in improving robustness against data quality issues, such as reduced training data and image degradations.

Our main contributions are:
\begin{itemize}
    \item We identify the underexplored potential of image augmentations in \neuralname and raise pertinent questions regarding their integration and impact on the learning process.

    \item We introduce and analyze two distinct image augmentation setups, SIA and DIA, for incorporating such augmentations into \neuralname, and study their performance in terms of photometric quality and surface reconstruction.

    \item We present a comprehensive empirical study that highlights the advantages of image augmentations in handling data quality issues, such as reduced training data and image degradations.
\end{itemize}

Overall, our results provide valuable insights into the potential of incorporating image augmentations into \neuralname, underlining the advantages of the simpler SIA setup with respect to photometric quality, surface reconstruction quality, and robustness against issues in data quality.
We provide our PyTorch~\cite{NEURIPS2019_9015} implementation at \url{https://github.com/juancprzs/ImAug_NeuralRendering}.

\section{Related Work}

\paragraph{Neural Rendering Methods.}
Recently, Neural Rendering Methods (\neuralname) have emerged as a powerful tool for representing 3D scenes and objects, as they can faithfully capture the appearance of complex structures~\cite{tewari2020state, tewari2022advances}.
Several approaches have been proposed to learn continuous representations of 3D scenes, including Signed Distance Functions (SDF)~\cite{park2019deepsdf}, occupancy networks~\cite{mescheder2019occupancy}, and Neural Radiance Fields~(NeRF)~\cite{mildenhall2021nerf}.
\neuralname, such as NeuS~\cite{wang2021neus}, NeRF, and NGP~\cite{mueller2022instant}, represent a scene as a continuous function learned by a deep neural network, allowing for geometry reconstruction and novel view synthesis with high fidelity.
Our work extends the literature on \neuralname by analyzing the integration of image augmentations into their learning process, and investigating how performance is affected in two distinct setups: Static Image Augmentations (SIA) and Dynamic Image Augmentations (DIA).

\paragraph{Image Augmentations in Computer Vision.}
Image augmentations have been widely used to improve the performance and robustness of various computer vision tasks, such as object recognition~\cite{resnet, Krizhevsky2012imagenet, simonyan2014very, perez2021enhancing}, object detection~\cite{li2017fully} and segmentation~\cite{moshkov2020test}.
Augmentations typically involve spatial transformations, like scaling, rotation, or flipping, as well as photometric transformations, such as enhancing contrast or manipulating brightness or hue.
While image augmentations are customarily employed across various applications in computer vision, their study within the context of \neuralname has been limited~\cite{wang2023benchmarking, mildenhall2022rawnerf, martinbrualla2020nerfw}.
In this work, we explore the challenges of incorporating image augmentations into the training of \neuralname and study the benefits these augmentations bring on performance and robustness.

\paragraph{Appearance Embeddings and Disentangling Geometry and Appearance.}
Disentangling the geometry and appearance of a scene can benefit its reconstruction~\cite{yariv2020multiview, yariv2021volume}.
A practical way to allow the model to explain away appearance variations is by conditioning the model on codes or embeddings that account for such discrepancies.
This technique has been successfully used to condition neural networks on specific appearance variations~\cite{karras2020analyzing, park2019SPADE, martinbrualla2020nerfw}.
In our work, we use appearance embeddings, following Brualla~\etal~\cite{martinbrualla2020nerfw}, to allow \neuralname to account for the photometric inconsistencies introduced by our image augmentations in both the SIA and DIA setups.

\section{Methodology}\label{sec:methodology}
In this section, we describe two approaches to introducing image augmentations into the training of neural rendering methods~(\neuralname): Static Image Augmentations (SIA) and Dynamic Image Augmentations (DIA).
Augmentations are ubiquitous in computer vision tasks, providing robustness to variations and improved generalization.
However, training \neuralname on augmented images is difficult because image augmentations may introduce geometric and photometric inconsistencies into the training data.
Correspondingly, we address these two issues by~(1)~using only geometry-preserving augmentations, and~(2)~introducing augmentation-dependent representations into the \neuralname model. 
In this section, we first explain how we modify both the dataset and the \neuralname to accommodate training-time augmentations for both SIA and DIA. 
Finally, we compare the computational and memory requirements of SIA and DIA, and discuss their implications on the training of \neuralname.

\subsection{Augmented Datasets}\label{sec:data_aug}
\paragraph{Augmentations.}
Given a training dataset $\mathcal{D}$ of posed images, we preserve geometry by augmenting the images only with \textit{color manipulations}, since color changes affect the scene's appearance but not its geometric structures nor the camera's view of the scene.
Formally, we augment $\mathcal{D}$ with a set $\mathcal{M} = \{M_i\}_{i=1}^{N+1}$ of color manipulations. Specifically, in our implementation, we use $N=5$ color manipulations: contrast, hue, saturation, sharpness, and brightness.
We thus define the set of manipulations to contain these $N$ manipulations plus the ``identity'' manipulation, \ie $\mathcal{M} = \{M_{\text{cont}}, M_{\text{hue}}, M_{\text{sat}}, M_{\text{sharp}}, M_{\text{bright}}, M_{\text{id}}\}$, and thus $|\mathcal{M}| = N+1$. The intensity of each parameterized manipulation is controlled by a scalar $p$, such that an augmented image $I^\prime$ can be obtained from the~$i^{\text{th}}$ manipulation function as $I^\prime = M_i(I,p)$.

\paragraph{Static Image Augmentation.} 
SIA generates a \textit{static} manipulated dataset $\mathcal{D}^\prime$ before starting training.
In this setup, SIA assumes each manipulation is coupled with a single intensity parameter, \ie there is a given set of intensity hyper-parameters $\mathcal{P} = \{p_i\}$ corresponding to the set of manipulations $\mathcal{M}$. 
Given both $\mathcal{M}$ and $\mathcal{P}$, SIA constructs $\mathcal{D}^\prime$ by manipulating $|\mathcal{M}|$ replicas of $\mathcal{D}$, that is, one for each color manipulation and one for the original dataset without augmentations (the ``identity'' replica). 
We keep track of the manipulation that was applied to each replica, and then training is conducted as usual by sampling images uniformly from the manipulated dataset $\mathcal{D}^\prime$.

\paragraph{Dynamic Image Augmentation.} 
In contrast to SIA, DIA creates a \textit{dynamic} manipulated dataset $\mathcal{D}^\prime$ for each training iteration.
In this setup, DIA assumes each manipulation is coupled with a parameter characterizing a distribution from which the manipulations' intensities will be sampled, \ie there is a given set of width hyper-parameters $\mathcal{W} = \{w_i\}$ corresponding to the set of manipulations $\mathcal{M}$. 
Given both $\mathcal{M}$ and $\mathcal{W}$, DIA dynamically generates $\mathcal{D}^\prime$ by applying color manipulations on-the-fly to training images.
In particular, after sampling an image $I$ from $\mathcal{D}$, DIA additionally samples a manipulation $M_i$ and a scalar $p$, and generates an augmented version of the image as $M_i(I, p)$.
Formally, DIA samples $M_i$ from the discrete uniform distribution over the set of manipulations, that is $M_i \sim \mathcal{U}\left(\mathcal{M} \setminus M_{\text{id}}\right)$.
Additionally, DIA samples $p$ from a zero-centered uniform distribution of width $w_i$, as $p \sim \mathcal{U}[-\nicefrac{w_i}{2},\nicefrac{w_i}{2}]$. 
We define the range of the intensity $p$ such that, when $p = 0$, all manipulations behave like the identity transform, that is $I = M_i(I,0),\: \forall\:i$.
Thus, this distribution for $p$ ensures that the expected value of the transform is the identity, irrespective of $w_i$.

\subsection{Modified Neural Rendering Methods}
\paragraph{Standard \neuralname.}
We consider \neuralname $f_\theta$, parameterized by $\theta$, which represent 3D scenes via a continuous volumetric representation. 
Given a 3D position $\mathbf{x}$ and a 2D view direction $\mathbf{d}$, the function $f_\theta$ predicts the RGB emitted color $\mathbf{c}$ and density $\sigma$, \ie formally, $f_\theta: (\mathbf{x}, \mathbf{d}) \mapsto (\mathbf{c}, \sigma)$. 
Given these encoded inputs, $f_\theta$ is often defined in two steps, each using one MLP:
\begin{align}
\sigma_\theta(\mathbf{x}), z(\mathbf{x}) &= \text{MLP}_1(\mathbf{x}), \label{eqn:MLP1} \\
\mathbf{c}_\theta(\mathbf{d}, \mathbf{x}) &= \text{MLP}_2(\mathbf{d}, z(\mathbf{x})) \label{eqn:MLP2}.
\end{align}
Here, $z(\mathbf{x})$ is a positional latent code passed from the first to the second MLP to inform it about position information.
While both inputs $\mathbf{x}$ and $\mathbf{d}$ are often represented via a positional encoding~\cite{mildenhall2021nerf, tancik2020fourfeat, yuce2022structured} that helps \neuralname model high-frequency functions, our notation omits this fact to avoid clutter.

The emitted color and density predicted by $f_\theta$ can then be used to synthesize an image $\hat{I}(\theta)$ from a specific point of view via volumetric rendering~\cite{mildenhall2021nerf}. 
This rendering is performed pixel-wise: for a pixel with associated camera ray $\mathbf{r}(t) = \mathbf{o} + t\cdot\mathbf{d}$, \ $t \geq 0$ (with camera origin $\mathbf{o}$ and direction $\mathbf{d}$), volumetric rendering computes the pixel's color by accumulating the colors and densities along $\mathbf{r}$ via
\begin{equation}\label{eq:vol_rendering}
C(\mathbf{r}) = \int_{t_n}^{t_f}T(t)\:\sigma_\theta\left(\mathbf{r}(t)\right)\:\mathbf{c}_\theta(\mathbf{d}, \mathbf{r}(t))\:\mathrm{d}t, \qquad T(t) = \exp\left(-\int_{t_n}^{t}\sigma_\theta\left(\mathbf{r}(s)\right)\:\mathrm{d}s\right),
\end{equation}
where $t_n$ and $t_f$ denote the near and far bounds of integration, and $T(t)$ is the transmittance accumulated along the ray.

Thus, an image $I$ in the training set $\mathcal{D}$ can be used as supervision for the synthesized render $\hat{I}(\theta)$, allowing $f_\theta$ to be learnt via backpropagation.
Specifically, in terms of $\mathcal{D}$, \neuralname can be learnt by searching for the parameters $\theta$ that optimize the loss function
\begin{align*}
\mathcal{L} = \sum_{I\in\mathcal{D}} \norm{\hat{I}(\theta) - I}_2^2. %
\end{align*}

Unfortunately, while the original dataset $\mathcal{D}$ provides consistent supervision for learning $f_\theta$, \ie its contents describe a plausible scene, the manipulated dataset $\mathcal{D}^\prime$ may not: for instance, after hue manipulation, a ship may look red in some images and yellow in others, as in Figure~\ref{fig:pull_fig}.
Namely, while color manipulations preserve the \textit{geometric consistency} of the original images, they destroy the \textit{photometric consistency}. 
This inconsistency requires us to modify $f_\theta$ to allow training on the manipulated dataset $\mathcal{D}^\prime$. 

\paragraph{Modified \neuralname.}
We enable $f_\theta$ to account for the photometric inconsistencies in the manipulated images by using \textit{learnable} appearance embeddings~\cite{martinbrualla2020nerfw}. 
Specifically, we introduce one such embedding for each color manipulation in $\mathcal{M}$.
Thus, in addition to the position and direction inputs, our modified $f_\theta^\prime$ receives an appearance embedding~(refer to Figure~\ref{fig:pull_fig} for a visual guide to this modification).
The appearance embedding informs $f_\theta^\prime$ of the manipulation that was performed on the image, \ie the tuple $(M, p)$.
Formally, we modify Equation~\eqref{eqn:MLP2} to incorporate the appearance embedding, and thus compute color as
\begin{align}
\mathbf{c}_\theta\left(\mathbf{d}, \mathbf{x}, M, p\right) &= \text{MLP}_2\left(\mathbf{d}, [z(\mathbf{x}), \mathbf{e}_M^p]\right),\label{eqn:mod_INR}
\end{align}
where $[\cdot]$ denotes concatenation, and $\mathbf{e}_M^p \in \mathbb{R}^d$ is the appearance embedding corresponding to the manipulation $M$ of intensity $p$. 
The computation of color via Equation~\eqref{eqn:mod_INR} is shared by both augmentation approaches (SIA and DIA).
The inner workings of these approaches differ only in how the embedding $\mathbf{e}_M^p$ is defined.
Let ``$\mathrm{Emb}$'' be a small table of $|\mathcal{M}|$ learnable embeddings, one for each manipulation.
Then, in SIA, the appearance embedding is simply defined as $\mathbf{e}_M^p = \mathrm{Emb}(M)$, since the sets $\mathcal{M}$ and $\mathcal{P}$ are coupled, and thus $p$ is redundant. 
In DIA, the appearance embedding is defined by concatenating the manipulation's embedding with the intensity, \ie as $\mathbf{e}_M^p = [\mathrm{Emb}(M), p]$.

Given our proposed changes, $f_\theta^\prime$ can be used to synthesize an image $\hat{I}(\theta, M, p)$ from a specific viewpoint with Equation~\eqref{eq:vol_rendering}, \ie via volumetric rendering.
That is, on one hand, SIA can now use $\mathcal{D}^\prime$ to learn $\theta$ by minimizing the loss function
\begin{align*}
\mathcal{L}_{\text{SIA}} = \sum_{(I, M, p)\in\mathcal{D}^\prime} \norm{\hat{I}(\theta, M, p) - I}_2^2. %
\end{align*}
On the other hand, DIA can learn $\theta$ by optimizing the stochastic loss function
\begin{align*}
\mathcal{L}_{\text{DIA}} = \mathbb{E}_{M,p}\left[ \sum_{I\in\mathcal{D}} \norm{\hat{I}(\theta, M, p) - M(I, p)}_2^2 \right], %
\end{align*}
where the expected value for $M$ and $p$ is taken over the distributions defined in Section~\ref{sec:data_aug}.
\footnote{We make a slight abuse of notation of the expected value, and omit the functional dependence that the random variable $p$ has on $M$.}

At test-time, for both SIA and DIA, we synthesize images without color manipulations by explicitly querying $f_\theta^\prime$ to do so.
In particular, we use the appearance embedding corresponding to the identity transformation along with its associated intensity parameter, \ie we query $\hat{I}(\theta, M = M_{\text{id}}, p = 0)$. 

\subsection{Comparison of SIA and DIA}
We now compare the computational and memory overheads introduced by SIA and DIA.
Both methods introduce a minimal memory overhead due to the appearance embeddings consisting of the compact embedding table ``$\mathrm{Emb}$'' with $(N+1)$ embeddings, each of dimension $d$. 
Furthermore, the second network (MLP$_2$), requires a small increase in the parameters of its first layer, since it now must process the appearance embeddings.
In terms of hyper-parameters, SIA requires the set $\mathcal{P}$ while DIA requires $\mathcal{W}$, both with a cardinality of $(N+1)$.

\paragraph{SIA-specific overheads.} 
SIA incurs a memory overhead, as the dataset augmentation creates $N$ times the amount of extra data. 
However, this strategy eliminates the need for on-the-fly transformations during training.
That is, essentially all of SIA's overheads occur before training, and so does not scale with the training procedure's duration.

\paragraph{DIA-specific overheads.} 
DIA introduces essentially no memory overhead, but increases computation, as transformations must be applied on-the-fly to each sampled image.

In summary, SIA incurs a memory overhead due to the creation of $N$ times extra data, while DIA increases computation by applying transformations on-the-fly. 
The training data is expected to be more diverse under DIA, as each sampled transformation results in a stochastic version of the dataset.

\section{Experiments}
In this section, we first describe our experimental setup and implementation details.
Then, we present extensive experiments that assess how training Neural Rendering Methods (\neuralname) with color manipulations affects~(1)~photometric quality,~(2)~surface reconstruction quality, and~(3)~robustness against issues in data quality.

\subsection{Experimental Setup}
Across all experiments, we assess the performance of \neuralname along two main axes: render quality and geometric reconstruction quality. 

\paragraph{Assessing photometric quality.}
As study subjects for photometric quality, we consider two methods designed for synthesizing novel views: \emph{Neural Radiance Fields (NeRF)}~\cite{mildenhall2021nerf} and \emph{Instant Neural Graphics Primitives (NGP)}~\cite{mueller2022instant}.
We measure the photometric quality of the synthesized renders via PSNR, LPIPS~\cite{LPIPS}, and SSIM~\cite{SSIM}.
We report PSNR in the main paper, and leave LPIPS and SSIM to the \underline{Appendix}.
Correspondingly, we evaluate these novel-view synthesis methods in the \emph{Blender synthetic}~\cite{mildenhall2021nerf} dataset.
This dataset consists of eight synthetic scenes with non-Lambertian materials and intricate geometries. 
Each scene provides 400 posed images, divided into 100 views for training, 100 for validation and 200 for testing.
All images have a resolution of $800 \times 800$.

\paragraph{Assessing surface reconstruction quality.}
As study subject for surface reconstruction quality, we use \emph{NeuS}~\cite{wang2021neus}, a neural surface reconstruction method.
NeuS implicitly learns a signed distance function (SDF).
Thus, NeuS allows for easy extraction of the learned mesh by running Marching Cubes over the zero-level set of its SDF.
We assess the quality of the resulting geometric reconstruction via Chamfer distance between the predicted mesh and a ground-truth point cloud.
Correspondingly, we evaluate NeuS on the \emph{Multi View Stereo DTU}~\cite{jensen2014large} dataset, consisting of $1600 \times 1200$ real images.
Following standard practice~\cite{yariv2020multiview, viscogrids2022pumarola}, we evaluate our method on 15 scenes, each comprising either 49 or 64 images. 
All scenes were captured with fixed camera and lighting parameters, and offer ground-truth point clouds.

\subsection{Implementation Details}\label{sec:impl_details}
\paragraph{Codebase.}
We use fast implementations of all methods to allow for a comprehensive study.
Specifically, for NeRF and NGP, we use Nerfacc, by Li~\etal~\cite{li2023nerfacc}, while for NeuS we use the implementation of  Guo~\etal~\cite{instant_nsr_pl}.
The evaluation of Chamfer distance on DTU follows the implementation in~\cite{DTUeval}, with the variation that, for compatibility with other frameworks, we do not consider the Chamfer distance as the \textit{average} between the two Chamfer distances, but rather the \textit{sum}.

\paragraph{Training details.}
We train NGP for 20\texttt{k} iterations and both NeRF and NeuS for 50\texttt{k} iterations.
All methods are trained with Adam~\cite{kingma2014adam}, following the official configurations~\cite{li2023nerfacc, instant_nsr_pl}.
Both NGP and NeuS use a learning rate of $10^{-2}$, while NeRF uses a learning rate of $5\times10^{-4}$.
We run all of our experiments on an NVIDIA A100.
In this setup, training times w.r.t. the baseline (\ie no augmentations) are essentially preserved when introducing SIA.
However, introducing DIA induces an increase in training time of about 3$\times$ for NeRF and NeuS and 5$\times$ for NGP.
We report further details in the \underline{Appendix}.

\paragraph{Repeated measurements.}
All results reported in the paper are the average across at least three runs.
Please refer to the \underline{Appendix} for the standard deviation of all reported measurements.

\subsection{Photometric Quality}\label{subsec:photom_qual}
We train NeRF and NGP on the Blender dataset with \textit{no} augmentations (\ie the baseline), our proposed dynamic augmentations (DIA), and our proposed static augmentations (SIA).

We report the PSNR values resulting from this experiment in Table~\ref{tab:PSNR_blender}.
We find that, for NeRF, DIA's performance is comparable to the baseline, achieving an average PSNR of 31.52 \textit{vs.} 31.54 of the baseline.
In contrast, SIA outperforms the baseline on seven of the eight scenes.
That is, SIA demonstrates that it consistently boosts NeRF's photometric quality, resulting in across-the-board improvements in performance.
Similarly, for NGP, Table~\ref{tab:PSNR_blender} shows that, while DIA maintains the baseline's performance (PSNR of 32.46), SIA provides consistent and sizable improvements.
In particular, equipping NGP with SIA boosts performance from 32.46 to 32.71, while introducing negligible computational overhead.

\begin{table}[t]
\centering
\setlength{\tabcolsep}{6pt}
\begin{tabular}{ll|cccccccc|l}
\hline  \toprule
\multirow{2}{*}{Model}  & \multirow{2}{*}{Aug.} & \multicolumn{8}{c|}{Scene}                                                                    & \multirow{2}{*}{Avg.} \\
                        &                       & Chair     & Drums     & Ficus             & Hotdog    & Lego      & Mater.    & Mic       & Ship      &                       \\
\hline
\multirow{3}{*}{NeRF}   & $-$                   & 33.12     & 25.40     & \textbf{32.68}    & 35.75     & \textit{33.65}     & \textit{29.66} &           \textit{33.81} &           \textit{28.23} &           \textit{31.54} \\
& DIA &           \textit{33.16} &           \textit{25.48} &           \textit{32.65} &           \textit{35.89} &           33.43 &           29.63 &           33.73 &           28.15 &           31.52 \\
& SIA &  \textbf{33.26} &  \textbf{25.55} &           32.61 &  \textbf{36.09} &  \textbf{33.67} &  \textbf{29.72} &  \textbf{33.83} &  \textbf{28.28} &  \textbf{31.63} \\
\hline
\multirow{3}{*}{NGP} & $-$ &  \textbf{35.17} &           \textit{25.18} &           31.63 &  \textbf{37.18} &  \textbf{35.54} &           \textit{29.06} &  \textbf{36.08} &  \textbf{29.86} &           32.46 \\
& DIA &           34.83 &           25.04 &           \textit{33.77} &           36.60 &           35.12 &           28.91 &           35.67 &           \textit{29.72} &           32.46 \\
& SIA &           \textit{35.10} &  \textbf{25.81} &  \textbf{33.91} &           \textit{36.92} &           \textit{35.36} &  \textbf{29.11} &           \textit{35.99} &           29.47 &  \textbf{32.71} \\
\bottomrule\hline
\end{tabular}
\vspace{0.15cm}
\caption{
\textbf{Impact of image augmentations on render quality in the Blender dataset.}
We report average PSNR across five runs (\ie higher is better).
We experiment with NeRF and NGP and equip them with our proposed dynamic (DIA) or static (SIA) image augmentations.
Our measurements find that, while DIA maintains the baselines' performance, SIA consistently provides boosts in render quality.
We \textbf{embolden} the best result and \textit{italicize} the runner-up.
}
\label{tab:PSNR_blender}
\end{table}

\subsection{Surface Reconstruction Quality}\label{subsec:surf_rec}
We train NeuS on the DTU dataset and introduce either no augmentations (the baseline), DIA, or SIA, and report results in Table~\ref{tab:Chamfer_DTU}.

With respect to surface reconstruction quality, we find that both DIA and SIA provide consistent improvements.
In particular, Table~\ref{tab:Chamfer_DTU} shows that the baseline is the \textit{worst} performer in nine of the 15 scenes evaluated.
We also observe that, while DIA outperforms the baseline in nine scenes, its average across the scenes is tied with that of the baseline's~(Chamfer of 2.03).
Finally, we note that \textit{SIA is the best performer across all setups in a total of 10 scenes}.
Correspondingly, this improved performance is reflected in its average performance across the scenes, where it improves the baseline's performance from 2.03 to 1.91.

\paragraph{SIA is cheaper and performs better than DIA.}
The improvements on photometric and surface reconstruction quality~(Tables~\ref{tab:PSNR_blender} and~\ref{tab:Chamfer_DTU}) show that image augmentations are a simple and intuitive approach to enhancing \neuralname.
In particular, our results find that static augmentations (SIA) provide sizable and consistent improvements over relevant baselines and across scenes.
Additionally, we underscore that SIA achieves these milestones while introducing virtually no computational overhead to the baseline.
Thus, \underline{the rest of our paper focuses on SIA}, where we simply refer to it as ``image augmentations'' (IA).

\begin{table}[t]
\centering
\setlength{\tabcolsep}{2.8pt}
\begin{tabular}{l|ccccccccccccccc|l}
\hline\toprule
\multirow{2}{*}{Aug.}   &           \multicolumn{15}{c|}{Scene}  &             \multirow{2}{*}{Avg.} \\
 &          24 &          37 &          40 &          55 &          63 &          65 &          69 &          83 &          97 &         105 &         106 &         110 &         114 &         118 &         122 &              \\
\midrule
$-$ &   \textbf{2.11} &            2.85 &            1.41 &   \textbf{0.84} &   \textbf{2.17} &            2.07 &   \textbf{2.21} &            2.72 &            3.17 &            1.61 &            \textit{1.51} &            4.17 &            \textit{0.85} &            1.58 &            1.24 &            2.03 \\
     DIA &            2.50 &            \textit{2.52} &            \textit{1.40} &            0.86 &            2.75 &            \textit{1.78} &            2.26 &            \textit{2.68} &            \textit{2.84} &            \textit{1.57} &            1.73 &   \textbf{3.90} &            0.93 &            \textit{1.50} &            \textit{1.16} &            2.03 \\
     SIA &            \textit{2.26} &   \textbf{2.15} &   \textbf{1.18} &            0.86 &            \textit{2.43} &   \textbf{1.66} &            \textit{2.24} &   \textbf{2.59} &   \textbf{2.79} &   \textbf{1.52} &   \textbf{1.46} &            \textit{4.12} &   \textbf{0.82} &   \textbf{1.48} &   \textbf{1.13} &   \textbf{1.91} \\
\bottomrule\hline
\end{tabular}
\vspace{0.15cm}
\caption{
\textbf{Impact of image augmentations on surface reconstruction quality in the DTU dataset.}
We report average Chamfer distances across five runs (\ie lower is better).
We equip NeuS with either dynamic (DIA) or static (SIA) image augmentations.
Our measurements find that DIA improves the baseline's performance in the majority of scenes.
Furthermore, we find that SIA is the best performer, consistently lowering the reconstruction error in 10 scenes out of 15 and improving the average error.
We \textbf{embolden} the best result and \textit{italicize} the runner-up.
}
\label{tab:Chamfer_DTU}
\end{table}

\subsection{Robustness Against Data Deficiencies}\label{sec:robustness}
Having observed how \neuralname benefit from Image Augmentations (IA) in standard setups, we turn to study the effects of IA when \neuralname are exposed to deficient training data.
In particular, we study two ways in which training data may present quality issues:~(1)~reduced number of training images, and~(2)~image degradations.

For studying the setup with a reduced number of images, we randomly subsample the training datasets according to a target percentage, where we vary this percentage in~$\{10, 25, 50, 75\}$.
For the case of image degradations, we consider five types of image degradations that commonly occur when capturing images in the wild.
In particular, we consider motion blur plus four types of camera sensor noise: Gaussian, Poisson, Salt and Pepper (S\&P), and Speckle.
We model each of these degradations with a single parameter.
Please refer to the \underline{Appendix} for details of how these degradations are modeled, the parameters used in this experiment, and additional results with other parameter values.

\begin{table}[t]
\centering
\setlength{\tabcolsep}{5pt}
\begin{tabular}{ll|cccc||ccccc}
\hline\toprule
\multirow{2}{*}{Method}  & \multirow{2}{*}{IA}  & \multicolumn{4}{c||}{Reduced dataset (\%)} & \multicolumn{5}{c}{Noise} \\
    &       &              10 &              25 &              50 &              75 &           Gaussian &     Motion &         Poisson &      S\&P &         Speckle \\
\midrule
\multirow{2}{*}{NeRF} & \xmark &           24.76 &           28.71 &           30.75 &           31.28 &           26.90 &           27.63 &           26.33 &           24.66 &           24.98 \\
    & \cmark  &  \textbf{25.12} &  \textbf{28.92} &  \textbf{30.88} &  \textbf{31.37} &  \textbf{27.03} &  \textbf{27.69} &  \textbf{26.37} &  \textbf{24.74} &  \textbf{25.02} \\
\cline{1-11}
\multirow{2}{*}{NGP} & \xmark &           20.11 &           27.18 &           30.57 &           31.75 &           24.18 &           27.17 &           25.42 &           23.27 &           24.01 \\
    & \cmark  &  \textbf{21.26} &  \textbf{28.06} &  \textbf{31.36} &  \textbf{32.21} &  \textbf{24.41} &  \textbf{27.20} &  \textbf{25.47} &  \textbf{24.43} &  \textbf{24.05} \\
\bottomrule\hline
\end{tabular}
\vspace{0.15cm}
\caption{
\textbf{Effects of deficient training data on photometric quality.}
We report average PSNR across three runs~(higher is better).
We perturb the Blender dataset by introducing two types of data quality issues: reduced number of training images (left), and image degradations (right).
We train both NeRF and NGP on this perturbed dataset, and find that introducing Image Augmentations~(IA) provides across-the-board improvements for both methods and against all the types of data quality issues we explored.
Conventions: ``Motion'' is motion blur, and ``S\&P'' is Salt \& Pepper noise.
}
\label{tab:psnr_blender_perts}
\end{table}

\begin{figure}[t]
 \centering
 \includegraphics[width=1.\textwidth]{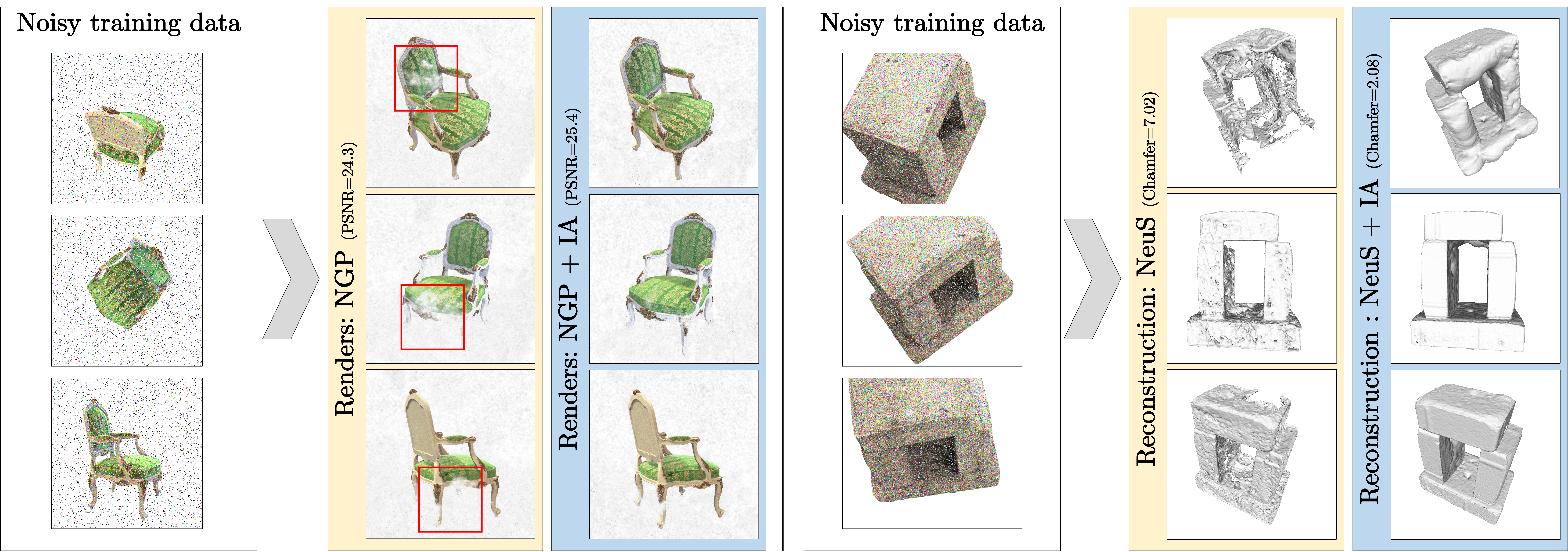}
 \caption{
     \textbf{Robustness against issues in data quality.}
     We introduce camera noise in the training data of \neuralname and compare their performance when equipped with (static) Image Augmentations (IA). 
     Left: when exposed to Salt \& Pepper noise, NGP's photometric quality improves with IA by reducing floating artifacts around the chair.
     Right: when exposed to speckle noise, IA dramatically improve NeuS' surface reconstruction quality (Chamfer distance drops from 7.02 to 2.08).
 }
 \label{fig:qualit_data_issues}
\end{figure}

\paragraph{Impacts on Photometric Quality.}
We first study the impact of data quality issues on photometric quality.
For this purpose, we artificially introduce these quality issues on the training data of Blender, and then train both NeRF and NGP on this dataset.
We report our results in terms of average PSNR on the Blender dataset in Table~\ref{tab:psnr_blender_perts}.
\underline{Reduced number of training images.}
The left section of Table~\ref{tab:psnr_blender_perts} shows that, when exposed to a reduced number of training images, all methods suffer in photometric quality.
We find that NGP, in particular, degrades faster than NeRF when given access to fewer images.
For instance, when operating on a reduced dataset of only 10\% of the images, NGP's original PSNR of 32.46 drops to 20.11 (a 12-point drop), while NeRF's original PSNR of 31.54 drops to 24.76 (a 7-point drop).
Our results also find that IA provides across-the-board improvements with respect to various training data sizes.
Specifically, \textit{introducing IA improves PSNR for both methods and for all reductions of the training dataset}.
\underline{Image degradations.}
The right section of Table~\ref{tab:psnr_blender_perts} demonstrates that common degradations in photographs have sizable impacts on photometric quality.
Overall, we again find that NeRF is more robust than NGP against data quality issues: Gaussian noise degrades NGP's performance to 24.41, while NeRF degrades more gracefully to 27.03.
Notably, we find that \textit{introducing IA consistently improves photometric quality for both methods and for all types of noise}.
We display some qualitative examples in the left side of Figure~\ref{fig:qualit_data_issues}, where we show that, when exposed to Salt \& Pepper noise, introducing IA improves the photometric quality of the renders (in this case improving PSNR by $\sim$1 point) by reducing floating artifacts.

\paragraph{Impacts on Surface Reconstruction Quality.}
We artificially introduce data quality issues on the DTU dataset and then train NeuS on this perturbed dataset.
We report our measurements of Chamfer distance in Table~\ref{tab:Chamfer_dtu_perts}.
\underline{Reduced number of training images.}
The left section of Table~\ref{tab:Chamfer_dtu_perts} demonstrates the impact of reducing the training set on surface reconstruction quality.
We observe that NeuS is mostly robust against the majority of reductions in the number of training instances.
For example, its original performance of 2.03 only degrades to 2.28 when having access to 50\% of the data.
However, when the training set is strongly reduced to 10\%, the performance of NeuS degrades dramatically to over 5.7.
Notably, we observe that \textit{introducing IA improves performance across all reduction percentages of the training dataset}.
For instance, when exposed to only 75\% of the data, NeuS' performance with IA is 2.02, which is still better than the baseline NeuS' original performance of 2.03 when exposed to 100\% of the data (see Table~\ref{tab:Chamfer_DTU}).
\underline{Image degradations.}
The right section of Table~\ref{tab:Chamfer_dtu_perts} shows that common photograph degradations have substantial effects on surface reconstruction quality.
For instance, speckle noise degrades the original performance in Chamfer distance from 2.03 to 4.54 (a 2.5-point increase).
Introducing IA makes performance degrade more gracefully, from 1.91 to only 3.91 (a 2-point increase).
Notably, our results show that \textit{equipping NeuS with IA consistently improves surface reconstruction quality against all types of noise}.
We further showcase some qualitative examples in the right side of Figure~\ref{fig:qualit_data_issues}.
Specifically, we observe that introducing speckle noise strongly degrades the performance of NeuS.
However, if NeuS is equipped with IA, its surface reconstruction improves dramatically (in this particular scene from 7.02 to 2.08).

\begin{table}[t]
\centering
\setlength{\tabcolsep}{8pt}
\begin{tabular}{l|cccc||ccccc}
\hline\toprule
\multirow{2}{*}{IA}  & \multicolumn{4}{c||}{Reduced dataset (\%)} & \multicolumn{5}{c}{Noise} \\
  &             10 &             25 &             50 &             75 &          Gaussian &    Motion &        Poisson &     S\&P &        Speckle \\
\midrule
\xmark &           5.72 &           2.91 &           2.28 &           2.15 &           2.36 &           2.01 &           3.34 &           2.68 &           4.54 \\
\cmark  &  \textbf{4.77} &  \textbf{2.65} &  \textbf{2.12} &  \textbf{2.02} &  \textbf{2.18} &  \textbf{1.90} &  \textbf{2.85} &  \textbf{2.39} &  \textbf{3.91} \\
\bottomrule\hline
\end{tabular}
\vspace{0.15cm}
\caption{
\textbf{Effects of deficient training data on surface reconstruction quality.}
We report average Chamfer distance across three runs~(lower is better).
We perturb the DTU dataset with two types of data quality issues (reduced number of training images, left, and image degradations, right), and train NeuS on this perturbed dataset.
We find that introducing Image Augmentations~(IA) provides consistent improvements against all the types of data quality issues we explored.
Conventions: ``Motion'' is motion blur, and ``S\&P'' is Salt \& Pepper noise.
}
\label{tab:Chamfer_dtu_perts}
\end{table}

Overall, our results show that Static Image Augmentations (SIA) stand as a reliable and simple way to improve both the photometric quality and surface reconstruction quality of \neuralname.
In particular, we have observed this finding to be consistent across relevant methods (NeRF, NGP, and NeuS) and datasets (Blender and DTU).
Furthermore, we have evidenced that these improvements in performance extend even to enhancing robustness against common issues in data quality, such as reduced training data and image degradations.

\section{Conclusions and Limitations}\label{sec:conclusion}
In this study, we have proposed and investigated two approaches for incorporating image augmentations into the training of \neuralname: Static Image Augmentations~(SIA) and Dynamic Image Augmentations~(DIA). 
Both methods focus on geometry-preserving augmentations, specifically color manipulations, in order to address the challenges arising from geometric and photometric inconsistencies introduced by augmentations. 
Through our methodologies, we demonstrated that the inclusion of image augmentations, in particular with the static approach~(SIA), into \neuralname contributes to improved performance and robustness.

Despite the promising results of our proposed approaches, certain limitations should be acknowledged. 
Both SIA and DIA focus on color manipulations as augmentations, which may not provide sufficient diversity and robustness for certain scenarios. 
Additionally, our methods rely on geometry-preserving augmentations, which limits the applicability to more complex transformations involving changes in shape or viewpoint. 
Furthermore, while DIA offers the advantage of a more diverse training dataset, the increased computational overhead may become a significant concern for large-scale training scenarios, limiting its usability.

In conclusion, our proposed methods, SIA and DIA, provide valuable insights into incorporating image augmentations into the training of \neuralname models. Future work may extend these techniques to include more complex and diverse augmentations, as well as explore alternative strategies to address the limitations related to memory and computational overheads. Overall, our work contributes to the advancement of neural rendering and its applications in computer vision, demonstrating the potential benefits of incorporating image augmentations into the training process.

{\small
\bibliography{egbib}
\bibliographystyle{ieee}
}

\clearpage
\appendix
\section{Appendix: \normalsize{\underline{Enhancing Neural Rendering Methods (\neuralname) with Image Augmentations}}}

\subsection{Render quality: SSIM and LPIPS}
In Sections~\ref{subsec:photom_qual} and~\ref{sec:robustness} we presented results of photometric quality in terms of PSNR~(Tables~\ref{tab:PSNR_blender} and~\ref{tab:Chamfer_dtu_perts}, in particular).
Here, we report additional results in photometric quality, analogous to such tables but in terms of the SSIM and LPIPS metrics.
Specifically, as presented here in the \underline{Appendix}, Tables~\ref{tab:SSIM_blender} and~\ref{tab:LPIPS_blender} are analogous to Table~\ref{tab:PSNR_blender} in the main paper (main results on photometric quality).
Furthermore, Tables~\ref{tab:ssim_blender_perts} and~\ref{tab:lpips_blender_perts} are analogous to Table~\ref{tab:psnr_blender_perts} in the main paper (results on robustness against deficiencies in data quality).
Broadly, both in terms of SSIM and LPIPS, our results here agree with the results in the main paper, as suggested by the PSNR values reported there.

\begin{table}[h]
\centering
\setlength{\tabcolsep}{6pt}
\begin{tabular}{ll|cccccccc|l}
\hline  \toprule
\multirow{2}{*}{Model}  & \multirow{2}{*}{Aug.} & \multicolumn{8}{c|}{Scene}                                                                    & \multirow{2}{*}{Avg.} \\
                        &                       & Chair     & Drums     & Ficus             & Hotdog    & Lego      & Mater.    & Mic       & Ship      &                       \\
\hline
\multirow{3}{*}{NeRF} & $-$ &           0.967 &           0.929 &           0.975 &           0.970 &           \textit{0.965} &           0.949 &           0.983 &           \textit{0.845} &           0.948 \\
& DIA &           0.967 &           \textit{0.930} &           0.975 &           \textit{0.971} &           0.963 &           0.949 &           0.983 &           0.844 &           0.948 \\
& SIA &  \textbf{0.969} &  \textbf{0.931} &  \textbf{0.976} &  \textbf{0.972} &  \textbf{0.965} &  \textbf{0.951} &  \textbf{0.983} &  \textbf{0.847} &  \textbf{0.949} \\
\hline
\multirow{3}{*}{NGP} & $-$ &  \textbf{0.983} &           0.927 &           0.976 &  \textbf{0.980} &  \textbf{0.979} &           0.940 &  \textbf{0.990} &  \textbf{0.885} &           0.957 \\
& DIA &           0.982 &           0.927 &           \textit{0.980} &           0.978 &           0.977 &           0.940 &           0.989 &           \textit{0.882} &           0.957 \\
& SIA &           \textit{0.983} &  \textbf{0.935} &  \textbf{0.980} &           \textit{0.979} &           \textit{0.978} &  \textbf{0.942} &           0.989 &           0.881 &  \textbf{0.958} \\
\bottomrule\hline
\end{tabular}
\vspace{0.15cm}
\caption{
\textbf{Impact of image augmentations on render quality (SSIM) in the Blender dataset.}
We report average SSIM across five runs (\ie higher is better).
This table is analogous to Table~\ref{tab:PSNR_blender}, but w.r.t. SSIM, instead of PSNR.
We \textbf{embolden} the best result and \textit{italicize} the runner-up.
}
\label{tab:SSIM_blender}
\end{table}

\begin{table}[h]
\centering
\setlength{\tabcolsep}{6pt}
\begin{tabular}{ll|cccccccc|l}
\hline  \toprule
\multirow{2}{*}{Model}  & \multirow{2}{*}{Aug.} & \multicolumn{8}{c|}{Scene}                                                                    & \multirow{2}{*}{Avg.} \\
                        &                       & Chair     & Drums     & Ficus             & Hotdog    & Lego      & Mater.    & Mic       & Ship      &                       \\
\hline
\multirow{3}{*}{NeRF} & $-$ &            0.038 &            0.073 &            0.026 &            0.046 &            \textit{0.041} &            \textit{0.052} &            0.019 &            \textit{0.194} &            0.061 \\
& DIA &            \textit{0.037} &            0.073 &            0.026 &            \textit{0.045} &            0.044 &            0.053 &   \textbf{0.019} &            0.196 &            0.061 \\
& SIA &   \textbf{0.036} &   \textbf{0.072} &   \textbf{0.026} &   \textbf{0.042} &   \textbf{0.040} &   \textbf{0.050} &            0.019 &   \textbf{0.191} &   \textbf{0.059} \\
\hline
\multirow{3}{*}{NGP} & $-$ &   \textbf{0.017} &            \textit{0.075} &            0.026 &   \textbf{0.029} &   \textbf{0.018} &            0.063 &   \textbf{0.012} &   \textbf{0.127} &            \textit{0.046} \\
& DIA &            0.018 &            0.076 &            \textit{0.024} &            0.031 &            0.020 &            0.064 &            0.013 &            \textit{0.131} &            0.047 \\
& SIA &            0.018 &   \textbf{0.066} &   \textbf{0.023} &            \textit{0.030} &            \textit{0.019} &   \textbf{0.060} &            0.013 &            0.132 &   \textbf{0.045} \\
\bottomrule\hline
\end{tabular}
\vspace{0.15cm}
\caption{
\textbf{Impact of image augmentations on render quality (LPIPS) in the Blender dataset.}
We report average LPIPS across five runs (\ie lower is better).
This table is analogous to Table~\ref{tab:PSNR_blender}, but w.r.t. LPIPS, instead of PSNR.
We \textbf{embolden} the best result and \textit{italicize} the runner-up.
}
\label{tab:LPIPS_blender}
\end{table}

\begin{table}[h]
\setlength{\tabcolsep}{5pt}
\begin{tabular}{ll|cccc||ccccc}
\hline\toprule
\multirow{2}{*}{Method}  & \multirow{2}{*}{IA}  & \multicolumn{4}{c||}{Reduced dataset (\%)} & \multicolumn{5}{c}{Noise} \\
    &       &              10 &              25 &              50 &              75 &           Gaussian &     Motion &         Poisson &      S\&P &         Speckle \\
\midrule
\multirow{2}{*}{NeRF} & \xmark &           0.894 &           0.931 &           0.944 &           0.947 &           0.912 &           0.921 &           0.943 &           0.888 &           0.916 \\
    & \cmark  &  \textbf{0.898} &  \textbf{0.933} &  \textbf{0.946} &  \textbf{0.948} &  \textbf{0.915} &  \textbf{0.923} &  \textbf{0.944} &  \textbf{0.889} &  \textbf{0.917} \\
\cline{1-11}
\multirow{2}{*}{NGP} & \xmark &           0.831 &           0.921 &           0.947 &           0.953 &  \textbf{0.900} &  \textbf{0.922} &           0.942 &           0.796 &  \textbf{0.859} \\
    & \cmark  &  \textbf{0.851} &  \textbf{0.928} &  \textbf{0.951} &  \textbf{0.956} &           0.898 &           0.922 &  \textbf{0.943} &  \textbf{0.797} &           0.858 \\
\bottomrule
\end{tabular}
\vspace{0.15cm}
\caption{
\textbf{Effects of deficient training data on photometric quality (SSIM).}
We report average SSIM across three runs~(higher is better).
We perturb the Blender dataset by introducing two types of data quality issues: reduced number of training images (left), and image degradations (right).
This table is analogous to Table~\ref{tab:psnr_blender_perts}, but w.r.t. SSIM, instead of PSNR.
Conventions: ``Motion'' is motion blur, and ``S\&P'' is Salt \& Pepper noise.
}
\label{tab:ssim_blender_perts}
\end{table}

\begin{table}[h]
\setlength{\tabcolsep}{5pt}
\begin{tabular}{ll|cccc||ccccc}
\hline\toprule
\multirow{2}{*}{Method}  & \multirow{2}{*}{IA}  & \multicolumn{4}{c||}{Reduced dataset (\%)} & \multicolumn{5}{c}{Noise} \\
    &       &              10 &              25 &              50 &              75 &           Gaussian &     Motion &         Poisson &      S\&P &         Speckle \\
\midrule
\multirow{2}{*}{NeRF} & \xmark &           0.105 &           0.073 &           0.064 &           0.062 &           0.170 &           0.083 &           0.064 &           0.219 &           0.080 \\
    & \cmark  &  \textbf{0.100} &  \textbf{0.070} &  \textbf{0.062} &  \textbf{0.060} &  \textbf{0.166} &  \textbf{0.081} &  \textbf{0.062} &  \textbf{0.218} &  \textbf{0.078} \\
\cline{1-11}
\multirow{2}{*}{NGP} & \xmark &           0.175 &           0.081 &           0.056 &           0.049 &  \textbf{0.244} &  \textbf{0.080} &           0.068 &           0.410 &           0.132 \\
    & \cmark  &  \textbf{0.154} &  \textbf{0.072} &  \textbf{0.051} &  \textbf{0.047} &           0.249 &           0.082 &  \textbf{0.068} &  \textbf{0.390} &  \textbf{0.131} \\
\bottomrule
\end{tabular}
\vspace{0.15cm}
\caption{
\textbf{Effects of deficient training data on photometric quality (LPIPS).}
We report average LPIPS across three runs~(lower is better).
We perturb the Blender dataset by introducing two types of data quality issues: reduced number of training images (left), and image degradations (right).
This table is analogous to Table~\ref{tab:psnr_blender_perts}, but w.r.t. LPIPS, instead of PSNR.
Conventions: ``Motion'' is motion blur, and ``S\&P'' is Salt \& Pepper noise.
}
\label{tab:lpips_blender_perts}
\end{table}

\subsection{Training times}
As mentioned in the main paper, we run all of our experiments in an NVIDIA A100, training NeRF for 50\texttt{k} iterations, NGP for 20\texttt{k}, and NeuS for 50\texttt{k}, which correspond to the defaults in the repositories of their respective implementations.
Given these numbers of iterations, the time for training the baselines~(\ie no augmentations) in minutes is: NeRF $\sim$35, NGP $\sim$6, and NeuS $\sim$28.
\textit{Note:} for reference and fair comparison among methods, if all baselines were trained for a standard of 20\texttt{k} iterations, these times would~(approximately) correspond to: NeRF $\sim$14, NGP $\sim$6, and NeuS $\sim$11.

When introducing SIA, training times are essentially preserved: NeRF $\sim$36 minutes, NGP $\sim$6 minutes, and NeuS $\sim$28 minutes.
However, when introducing DIA, the changes in training time are more dramatic, resulting in: NeRF $\sim$90 minutes, NGP $\sim$29 minutes, and NeuS $\sim$84 minutes.
These absolute changes in training time correspond to relative changes, w.r.t. the baselines, of $\sim$3$\times$ for NeRF~(35$\to$90), $\sim$5$\times$ for NGP~(6$\to$29), and $\sim$3$\times$ for NeuS~(28$\to$84).
In summary, introducing DIA induces an increase of about 6$\times$ in training time for all methods.

\subsection{Implementation of image degradations}\label{sec:impl_img_deg}
In Section~\ref{sec:robustness}, we introduced five image degradations into the training set of \neuralname.
Each degradation was adapted to depend on a single intensity parameter $q$.
Here, we elaborate on how these image degradations were modeled and the parameters that were used.

All images were codified to have their pixels lie in $[0, 1]$ for the degradations.
After degrading a pixel intensity $x$ into $x'$, we ensured the range of the resulting images stayed within the natural bounds via a clipping operation, \ie $x' \coloneqq \text{clip}(x', 0, 1)$.

\paragraph{Gaussian Noise.}
We introduced pixel-wise Gaussian noise.
A pixel with intensity $x$ is perturbed via $x' = x + \epsilon$, where $\epsilon \sim \mathcal{N}(0, \sigma^2)$.
The degradation's parameter corresponds to the Gaussian's standard deviation, where use $\sigma = q = 0.1$.

\paragraph{Motion Blur.}
A motion blur effect was introduced to images using a normalized line-shaped blur kernel, where the orientation and extent of the motion blur were defined by angle $\phi$ and kernel size $k$ respectively. 
The kernel was generated by setting a diagonal line of ones from the kernel's center in a direction defined by the given angle, $\phi$.
We modeled \textit{horizontal} motion blur, \ie $\phi=0$, and the kernel was independently applied to each image channel.
The degradation's parameter corresponds to the line's length (\ie the kernel size), for which we used $k = q = 5$.

\paragraph{Poisson Noise.}
Poisson noise was introduced to the images on a pixel-wise basis. 
This noise type is parameterized by a scale parameter, $\lambda$, stating the noise's variance. 
For a given pixel with intensity $x$, the noisy pixel value was modeled as a random variable $X$ drawn from a Poisson distribution, $X \sim \text{Poisson}(\lambda \cdot x)$. 
The resultant noisy pixel values were then normalized by dividing by the scale parameter, yielding $x' = X / \lambda$. 
In this context, the degradation's parameter corresponds to the scale, which we set to $\lambda = q = 10$.\footnote{Given our implementation of Poisson noise, this is the only noise for which a \textit{larger} parameter $q$ actually means a \textit{lower} intensity of the noise.}

\paragraph{Salt and Pepper Noise.}
Impulse noise, also known as salt and pepper noise, was introduced to the images. 
This noise randomly assigns the maximum value (salt) or minimum value (pepper) to selected pixels. 
For a given pixel with intensity $x$, the noisy pixel value $x'$ was modeled as a random variable taking values 0 or 1 according to two Bernoulli distributions, \ie, $x' \sim \text{Bernoulli}(p_{\text{salt}})$ for salt noise (with $x' = 1$) and $x' \sim \text{Bernoulli}(p_{\text{pepper}})$ for pepper noise (with $x' = 0$). 
To control the degradation with a single parameter, we assumed equal probabilities for both salt and pepper noise, and so used $p_{\text{salt}} = p_{\text{pepper}} = q = 0.05$.

\paragraph{Speckle Noise.}
Finally, we introduced speckle noise to the images. 
Speckle noise is a multiplicative noise that is often observed in radar imagery. 
Here we modeled the multiplicative component with a Gaussian distribution.
In particular, we model speckle noise as perturbing a pixel with intensity $x$ into $x' = x + x \cdot \mathcal{N}(0, \sigma^2)$. 
The degradation's parameter corresponds to $\sigma = q = 0.4$.

\subsection{Additional image degradation intensities}
In Section~\ref{sec:robustness}, we exposed methods to five types of image degradations, and recorded their changes in performance~(both in terms of photometric quality and surface reconstruction quality).
As explained in Section~\ref{sec:impl_img_deg}, each degradation was adapted to depend on a single intensity parameter $q$, whose magnitude was also reported in that section.
Here we report additional results on this experimental setup, where we \textit{lower} the degradation's intensity.
In particular, we report the impact of an additional intensity of image degradations in photometric quality and surface reconstruction quality in Tables~\ref{tab:more_intens_psnr}~and~\ref{tab:more_intens_chamfer}, respectively.

\begin{table}[ht]
\centering
\begin{tabular}{lll|cc}
\hline
\toprule
\multirow{2}{*}{Noise}   & \multirow{2}{*}{Noise parameter ($q$)} & \multirow{2}{*}{IA} &      \multicolumn{2}{c}{Method}                 \\
 & &  &        NeRF        &           NGP     \\
\midrule
\multirow{4}{*}{Gaussian} & \multirow{2}{*}{0.05} & \xmark &           29.16 &           27.68 \\
        &         & \cmark  &  \textbf{29.28} &  \textbf{29.83} \\
\cline{2-5}
        & \multirow{2}{*}{0.10} & \xmark &           26.90 &           24.18 \\
        &         & \cmark  &  \textbf{27.03} &  \textbf{24.41} \\
\cline{1-5}
\cline{2-5}
\multirow{4}{*}{Motion blur} & \multirow{2}{*}{2.5} & \xmark &  \textbf{25.54} &           24.91 \\
        &         & \cmark  &           \textbf{25.54} &  \textbf{25.04} \\
\cline{2-5}
        & \multirow{2}{*}{5.0} & \xmark &           27.63 &           27.17 \\
        &         & \cmark  &  \textbf{27.69} &  \textbf{27.20} \\
\cline{1-5}
\cline{2-5}
\multirow{4}{*}{Poisson} & \multirow{2}{*}{10} & \xmark &           26.33 &           25.42 \\
        &         & \cmark  &  \textbf{26.37} &  \textbf{25.47} \\
\cline{2-5}
        & \multirow{2}{*}{100} & \xmark &           30.50 &           30.71 \\
        &         & \cmark  &  \textbf{30.57} &  \textbf{30.86} \\
\cline{1-5}
\cline{2-5}
\multirow{4}{*}{Salt \& pepper} & \multirow{2}{*}{0.025} & \xmark &           27.85 &           27.16 \\
        &         & \cmark  &  \textbf{28.00} &  \textbf{28.25} \\
\cline{2-5}
        & \multirow{2}{*}{0.050} & \xmark &           24.72 &           23.27 \\
        &         & \cmark  &  \textbf{24.74} &  \textbf{24.43} \\
\cline{1-5}
\cline{2-5}
\multirow{4}{*}{Speckle} & \multirow{2}{*}{0.2} & \xmark &           28.57 &           28.10 \\
        &         & \cmark  &  \textbf{28.64} &  \textbf{28.25} \\
\cline{2-5}
        & \multirow{2}{*}{0.4} & \xmark &           24.98 &           24.01 \\
        &         & \cmark  &  \textbf{25.02} &  \textbf{24.05} \\
\bottomrule
\hline
\end{tabular}
\vspace{0.2cm}
\caption{
\textbf{Effects of various intensities of image degradations on photometric quality.}
We report average PSNR across three runs~(higher is better).
We perturb the Blender dataset with image degradations and train NeRF and NGP on this perturbed dataset.
Image Augmentations~(IA) provide consistent improvements against all types of noise and intensities we explored.
}
\label{tab:more_intens_psnr}
\end{table}

\begin{table}[h]
\centering
\begin{tabular}{lll|c}
\hline
\toprule
 Noise & Noise parameter ($q$) & IA &        NeuS    \\
\midrule
\multirow{4}{*}{Gaussian} & \multirow{2}{*}{0.05} & \xmark &           2.21 \\
        &         & \cmark  &  \textbf{2.07} \\
\cline{2-4}
        & \multirow{2}{*}{0.10} & \xmark &           2.36 \\
        &         & \cmark  &  \textbf{2.18} \\
\cline{1-4}
\cline{2-4}
\multirow{4}{*}{Motion blur} & \multirow{2}{*}{2.5} & \xmark &           2.19 \\
        &         & \cmark  &  \textbf{2.03} \\
\cline{2-4}
        & \multirow{2}{*}{5.0} & \xmark &           2.01 \\
        &         & \cmark  &  \textbf{1.90} \\
\cline{1-4}
\cline{2-4}
\multirow{4}{*}{Poisson} & \multirow{2}{*}{10} & \xmark &           3.34 \\
        &         & \cmark  &  \textbf{2.85} \\
\cline{2-4}
        & \multirow{2}{*}{100} & \xmark &           2.17 \\
        &         & \cmark  &  \textbf{2.04} \\
\cline{1-4}
\cline{2-4}
\multirow{4}{*}{Salt \& pepper} & \multirow{2}{*}{0.025} & \xmark &           2.43 \\
        &         & \cmark  &  \textbf{2.26} \\
\cline{2-4}
        & \multirow{2}{*}{0.050} & \xmark &           2.68 \\
        &         & \cmark  &  \textbf{2.39} \\
\cline{1-4}
\cline{2-4}
\multirow{4}{*}{Speckle} & \multirow{2}{*}{0.2} & \xmark &           2.40 \\
        &         & \cmark  &  \textbf{2.20} \\
\cline{2-4}
        & \multirow{2}{*}{0.4} & \xmark &           4.54 \\
        &         & \cmark  &  \textbf{3.91} \\
\bottomrule
\hline
\end{tabular}
\vspace{0.2cm}
\caption{
\textbf{Effects of various intensities of image degradations on surface reconstruction quality.}
We report average Chamfer distance across three runs~(lower is better).
We perturb the DTU dataset with image degradations and train NeuS on this perturbed dataset.
Image Augmentations~(IA) provide consistent improvements against all types of noise and intensities we explored.
}
\label{tab:more_intens_chamfer}
\end{table}

\clearpage
\subsection{Variation across runs}
As mentioned in Section~\ref{sec:impl_details}, all the results reported in the paper are the average across \textit{at least} three runs.
Here we report four tables (each one corresponding to one table in the paper), stating the standard deviation of the associated measurements.
In particular, we report Tables~\ref{tab:PSNR_blender_std},~\ref{tab:Chamfer_DTU_std},~\ref{tab:psnr_blender_perts_std}, and~\ref{tab:Chamfer_dtu_perts_std}, corresponding to Tables~\ref{tab:PSNR_blender},~\ref{tab:Chamfer_DTU},~\ref{tab:psnr_blender_perts}, and~\ref{tab:Chamfer_dtu_perts} of the main paper.
Additionally, Tables~\ref{tab:more_intens_chamfer_std}~and~\ref{tab:more_intens_psnr_std} display the standard deviations of the measurements reported in Tables~\ref{tab:more_intens_chamfer}~and~\ref{tab:more_intens_psnr}, here in the \underline{Appendix}.

\begin{table}[h]
\centering
\setlength{\tabcolsep}{6pt}
\begin{tabular}{ll|cccccccc|l}
\hline  \toprule
\multirow{2}{*}{Model}   &  \multirow{2}{*}{Aug.}   &           \multicolumn{8}{c|}{Scene}  &             \multirow{2}{*}{Avg.} \\
   &     &           Chair &           Drums &           Ficus &          Hotdog &            Lego &       Mater. &             Mic &            Ship &              \\
\hline
\multirow{3}{*}{NeRF} & - &  0.05 &  0.06 &  0.06 &  0.11 &  0.06 &           0.02 &           0.03 &  0.04 &  0.05 \\
& DIA &           0.01 &           0.04 &           0.04 &           0.05 &           0.04 &  0.03 &           0.03 &           0.02 &           0.03 \\
& SIA &           0.03 &           0.03 &           0.04 &           0.05 &           0.04 &           0.02 &  0.05 &           0.03 &           0.03 \\
\hline
\multirow{3}{*}{NGP} & - &           0.01 &  0.16 &  0.29 &  0.04 &  0.04 &  0.08 &           0.05 &           0.03 &  0.09 \\
& DIA &           0.01 &           0.08 &           0.08 &           0.02 &           0.04 &           0.02 &  0.09 &  0.04 &           0.05 \\
& SIA &  0.05 &           0.03 &           0.03 &           0.04 &           0.03 &           0.04 &           0.03 &           0.01 &           0.03 \\
\bottomrule\hline %
\end{tabular}
\vspace{0.15cm}
\caption{
\textbf{Standard deviations of Table~\ref{tab:PSNR_blender}: Impact of image augmentations on render quality in the Blender dataset.}
We report standard deviation of PSNR across \textit{five} runs.
}
\label{tab:PSNR_blender_std}
\end{table}

\begin{table}[h]
\centering
\setlength{\tabcolsep}{2.8pt}
\begin{tabular}{l|ccccccccccccccc|l}
\hline\toprule
\multirow{2}{*}{Aug.}   &           \multicolumn{15}{c|}{Scene}  &             \multirow{2}{*}{Avg.} \\
 &          24 &          37 &          40 &          55 &          63 &          65 &          69 &          83 &          97 &         105 &         106 &         110 &         114 &         118 &         122 &              \\
\midrule
- &  0.11 &           0.09 &           0.03 &  0.06 &           0.76 &           0.07 &           0.04 &           0.39 &           0.04 &  1.05 &           0.03 &           0.06 &           0.05 &           0.05 &  0.10 &           0.20 \\
DIA &           0.05 &  0.34 &           0.04 &           0.03 &  2.15 &           0.04 &  0.06 &  1.65 &           0.34 &           0.18 &           0.03 &  0.62 &           0.05 &           0.32 &           0.06 &           0.40 \\
SIA &           0.05 &           0.09 &  0.06 &           0.03 &           0.97 &  0.46 &           0.06 &           0.64 &  0.36 &           0.73 &  0.14 &           0.43 &  0.20 &  2.35 &           0.04 &  0.44 \\
\bottomrule\hline
\end{tabular}
\vspace{0.15cm}
\caption{
\textbf{Standard deviations of Table~\ref{tab:Chamfer_DTU}: Impact of image augmentations on surface reconstruction quality in the DTU dataset.}
We report standard deviation of Chamfer distances across \textit{five} runs.
}
\label{tab:Chamfer_DTU_std}
\end{table}

\begin{table}[h]
\centering
\setlength{\tabcolsep}{5pt}
\begin{tabular}{ll|cccc||ccccc}
\hline\toprule
\multirow{2}{*}{Method}  & \multirow{2}{*}{IA}  & \multicolumn{4}{c||}{Reduced dataset (\%)} & \multicolumn{5}{c}{Noise} \\
    &       &              10 &              25 &              50 &              75 &           Gaussian &     Motion &         Poisson &      S\&P &         Speckle \\
\midrule
\multirow{2}{*}{NeRF} & \xmark &           3.62 &  3.45 &           3.29 &           3.31 &           1.78 &  2.57 &  2.41 &           1.35 &  2.46 \\
    & \cmark  &  3.62 &           3.42 &  3.33 &  3.34 &  1.86 &           2.54 &           2.40 &  1.42 &           2.45 \\
\cline{1-11}
\multirow{2}{*}{NGP} & \xmark &           3.70 &  4.10 &  4.32 &  4.18 &           2.26 &  2.84 &           2.03 &           1.29 &           2.10 \\
    & \cmark  &  4.05 &           3.68 &           3.68 &           3.72 &  2.52 &           2.69 &  2.11 &  1.84 &  2.15 \\
\bottomrule\hline
\end{tabular}
\vspace{0.15cm}
\caption{
\textbf{Standard deviations of Table~\ref{tab:psnr_blender_perts}: Effects of deficient training data on photometric quality.}
We report standard deviations of PSNR across \textit{three} runs.
}
\label{tab:psnr_blender_perts_std}
\end{table}

\begin{table}[H]
\centering
\setlength{\tabcolsep}{8pt}
\begin{tabular}{l|cccc||ccccc}
\hline\toprule
\multirow{2}{*}{IA}  & \multicolumn{4}{c||}{Reduced dataset (\%)} & \multicolumn{5}{c}{Noise} \\
  &             10 &             25 &             50 &             75 &          Gaussian &    Motion &        Poisson &     S\&P &        Speckle \\
\midrule
\xmark &           2.28 &           1.45 &           1.00 &           0.97 &           1.09 &  0.86 &           2.04 &           1.33 &           2.30 \\
\cmark  &  1.92 &  1.29 &  0.92 &  0.86 &  1.02 &           0.87 &  1.75 &  1.26 &  2.29 \\
\bottomrule\hline
\end{tabular}
\vspace{0.15cm}
\caption{
\textbf{Standard deviations of Table~\ref{tab:Chamfer_dtu_perts}: Effects of deficient training data on surface reconstruction quality.}
We report standard deviations of Chamfer distance across \textit{three} runs.
}
\label{tab:Chamfer_dtu_perts_std}
\end{table}

\clearpage

\begin{table}[ht]
\centering
\begin{tabular}{lll|cc}
\hline
\toprule
\multirow{2}{*}{Noise}   & \multirow{2}{*}{Noise parameter ($q$)} & \multirow{2}{*}{IA} &      \multicolumn{2}{c}{Method}                 \\
 & &  &        NeRF        &           NGP     \\
\midrule
\multirow{4}{*}{Gaussian} & \multirow{2}{*}{0.05} & \xmark &           2.40 &  3.86 \\
        &         & \cmark  &  2.56 &           3.11 \\
\cline{2-5}
        & \multirow{2}{*}{0.10} & \xmark &           1.78 &           2.26 \\
        &         & \cmark  &  1.86 &  2.52 \\
\cline{1-5}
\cline{2-5}
\multirow{4}{*}{Motion blur} & \multirow{2}{*}{2.5} & \xmark &  2.04 &  2.14 \\
        &         & \cmark  &           1.99 &           2.08 \\
\cline{2-5}
        & \multirow{2}{*}{5.0} & \xmark &  2.57 &  2.84 \\
        &         & \cmark  &           2.54 &           2.69 \\
\cline{1-5}
\cline{2-5}
\multirow{4}{*}{Poisson} & \multirow{2}{*}{10} & \xmark &  2.41 &           2.03 \\
        &         & \cmark  &           2.40 &  2.11 \\
\cline{2-5}
        & \multirow{2}{*}{100} & \xmark &           2.90 &  3.02 \\
        &         & \cmark  &  2.92 &           3.00 \\
\cline{1-5}
\cline{2-5}
\multirow{4}{*}{Salt \& pepper} & \multirow{2}{*}{0.025} & \xmark &           2.12 &  2.22 \\
        &         & \cmark  &  2.27 &           2.03 \\
\cline{2-5}
        & \multirow{2}{*}{0.050} & \xmark &           1.35 &           1.29 \\
        &         & \cmark  &  1.42 &  1.84 \\
\cline{1-5}
\cline{2-5}
\multirow{4}{*}{Speckle} & \multirow{2}{*}{0.2} & \xmark &           2.55 &           2.37 \\
        &         & \cmark  &  2.57 &  2.42 \\
\cline{2-5}
        & \multirow{2}{*}{0.4} & \xmark &  2.46 &           2.10 \\
        &         & \cmark  &           2.45 &  2.15 \\
\bottomrule
\hline
\end{tabular}
\vspace{0.2cm}
\caption{
\textbf{Standard deviations of Table~\ref{tab:more_intens_psnr}: Effects of various intensities of image degradations on photometric quality.}
We report standard deviations of PSNR distance across \textit{three} runs.
}
\label{tab:more_intens_psnr_std}
\end{table}

\begin{table}[H]
\centering
\begin{tabular}{lll|c}
\hline
\toprule
 Noise & Noise parameter ($q$) & IA &        NeuS    \\
\midrule
\multirow{4}{*}{Gaussian} & \multirow{2}{*}{0.05} & \xmark &           1.04 \\
        &         & \cmark  &  0.94 \\
\cline{2-4}
        & \multirow{2}{*}{0.10} & \xmark &           1.09 \\
        &         & \cmark  &  1.02 \\
\cline{1-4}
\cline{2-4}
\multirow{4}{*}{Motion blur} & \multirow{2}{*}{2.5} & \xmark &           0.87 \\
        &         & \cmark  &  0.83 \\
\cline{2-4}
        & \multirow{2}{*}{5.0} & \xmark &  0.86 \\
        &         & \cmark  &           0.87 \\
\cline{1-4}
\cline{2-4}
\multirow{4}{*}{Poisson} & \multirow{2}{*}{10} & \xmark &           2.04 \\
        &         & \cmark  &  1.75 \\
\cline{2-4}
        & \multirow{2}{*}{100} & \xmark &           1.01 \\
        &         & \cmark  &  0.96 \\
\cline{1-4}
\cline{2-4}
\multirow{4}{*}{Salt \& pepper} & \multirow{2}{*}{0.025} & \xmark &  1.21 \\
        &         & \cmark  &           1.23 \\
\cline{2-4}
        & \multirow{2}{*}{0.050} & \xmark &           1.33 \\
        &         & \cmark  &  1.26 \\
\cline{1-4}
\cline{2-4}
\multirow{4}{*}{Speckle} & \multirow{2}{*}{0.2} & \xmark &           1.20 \\
        &         & \cmark  &  1.08 \\
\cline{2-4}
        & \multirow{2}{*}{0.4} & \xmark &           2.30 \\
        &         & \cmark  &  2.29 \\
\bottomrule
\hline
\end{tabular}
\vspace{0.2cm}
\caption{
\textbf{Standard deviations of Table~\ref{tab:more_intens_chamfer}: Effects of various intensities of image degradations on surface reconstruction quality.}
We report standard deviations of Chamfer distance distance across \textit{three} runs.
}
\label{tab:more_intens_chamfer_std}
\end{table}

\end{document}